# RH20T: A Comprehensive Robotic Dataset for Learning Diverse Skills in One-Shot

Hao-Shu Fang, Hongjie Fang, Zhenyu Tang, Jirong Liu, Chenxi Wang,
Junbo Wang, Haoyi Zhu and Cewu Lu
Shanghai Jiao Tong University
fhaoshu@gmail.com, {galaxies, tang_zhenyu, jirong, wcx1997, sjtuwjb3589635689, zhuhaoyi, lucewu}@sjtu.edu.cn

*Abstract*— A key challenge for robotic manipulation in open domains is how to acquire diverse and generalizable skills for robots. Recent progress in one-shot imitation learning and robotic foundation models have shown promise in transferring trained policies to new tasks based on demonstrations. This feature is attractive for enabling robots to acquire new skills and improve their manipulative ability. However, due to limitations in the training dataset, the current focus of the community has mainly been on simple cases, such as push or pick-place tasks, relying solely on visual guidance. In reality, there are many complex skills, some of which may even require both visual and tactile perception to solve. This paper aims to unlock the potential for an agent to generalize to hundreds of real-world skills with multi-modal perception. To achieve this, we have collected a dataset comprising over 110,000 *contact-rich* robot manipulation sequences across diverse skills, contexts, robots, and camera viewpoints, all collected *in the real world*. Each sequence in the dataset includes visual, force, audio, and action information. Moreover, we also provide a corresponding human demonstration video and a language description for each robot sequence. We have invested significant efforts in calibrating all the sensors and ensuring a high-quality dataset. The dataset is made publicly available on our website: rh20t.github.io.

## I. INTRODUCTION

Robotic manipulation requires the robot to control its actuator and change the environment following a task specification. Enabling robots to learn new skills with minimal effort is one of the ultimate goals of the robot learning community. Recent research in one-shot imitation learning [10, 14] and emerging foundation models [3, 5] draw an exciting picture of transferring trained policies to a new task given a demonstration. This paper shares the same aspiration.

While the future is promising, most research in robotics only demonstrates the effectiveness of their algorithms on simple cases, such as pushing, picking, and placing objects in the real world. Two main factors hinder the exploration of more complex tasks in this direction. Firstly, there is a lack of large and diverse robotic manipulation datasets in this field [3], despite the community's long-standing eagerness for such datasets. The fundamental problem stems from the huge barriers associated with data acquisition. These challenges include the arduous task of configuring diverse robot platforms, creating varied environments, and gathering manipulation trajectories, which require significant effort and resources. Secondly, most methods focus solely on visual guidance control, yet it has been observed in physiology that humans with impaired digital sensibility struggle to accomplish many daily manipulations with visual guidance alone [21]. This indicates that more sensory information should be considered in order to learn various manipulations in open environments.

To address these problems, we revisit the data collection process for robotic manipulation. In most imitation learning literature, expert robot trajectories are manually collected using simplified user interfaces like 3D mice, keyboards, or VR remotes. However, these control methods are inefficient and pose safety risks when the robot engages in rich-contact interactions with the environment. The main reasons are the unintuitive nature of controlling with a 3D mouse or keyboard, and the inaccuracies resulting from motion drifting when using a VR remote. Additionally, tele-operation without force feedback degrades manipulation efficiency for humans. In this paper, we equipped the robot with a force-torque sensor and employed a haptic device with force rendering for precise and efficient data collection. With the goal that the dataset should be representative, generalized, diverse and close to reality, we collect around 150 skills with complicated actions other than simple pick-place. These skills were either selected from RLBench [19] and MetaWorld [40], or proposed by ourselves. Many skills require the robot to engage in contact-rich interactions with the environment, such as cutting, plugging, slicing, pouring, folding, rotating, etc. We have used multiple different robot arms commonly found in labs worldwide to collect our dataset. The diversity in robot configurations can also aid algorithms in generalizing to other robots.

So far, we have collected around 110,000 sequences of robotic manipulation and 110,000 corresponding human demonstration videos for the same skills. This amounts to over 40 million frames of images for the robotic manipulation sequences and over 10 million frames for the human demonstrations. Each robot sequence contains abundant visual, tactile, audio, and proprioception information from multiple sensors. The dataset is carefully organized, and *we believe that a dataset with such diversity and scale is crucial for the future emergence of foundation models in general skill learning*, as promising progress has been witnessed in the NLP and CV communities [6, 32, 23].

## II. RELATED WORKS

We briefly review related works in robotic manipulation datasets, zero/one-shot imitation learning, and vision-force learning methods.

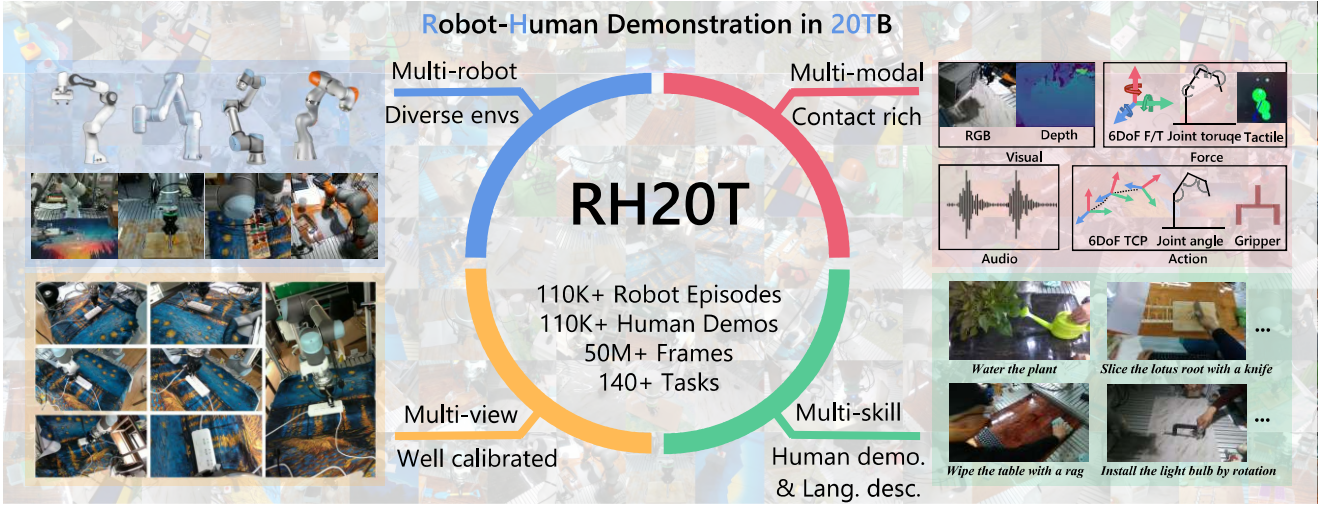

**Fig. 1:** Overview of our RH20T dataset. We adopt multiple robots and setup diverse environments for the data collection. The robot manipulation episodes include multi-modal visual, force, audio and action data. For each episode, we collect the manipulation process with well calibrated multi-view cameras. Our dataset contains diverse robotic manipulation skills and each episode has a corresponding human demonstration and language description. In total, we provide over 110K robot episodes and 110K corresponding human demonstration. The dataset contains over 50 million frames and over 140 tasks.

*a) Dataset:* Our community has been striving to create a large-scale and representative dataset for a significant period of time. Previous research in one-shot imitation learning has either collected robot manipulation data in the real world [14] or in simulation [27]. However, their datasets are usually small and the tasks are simple. Some attempts have been made to create large-scale real robot manipulation datasets [9, 15, 20, 22, 28, 34]. For example, RoboTurk [28] developed a crowd-sourcing platform and collected data on three tasks using mobile phone-based tele-operation. MIME [34] collected 20 types of manipulations using Baxter with kinesthetic teaching, but they were limited to a single robot and simple environments. RoboNet [9] gathered a significant amount of robot trajectories with various robots, grippers, and environments. However, it mainly consists of random walking episodes due to the challenges of performing meaningful skills. BC-Z [20] presents a manipulation collection of 100 "tasks", but as pointed out in [27], they are combinations of 9 verbs and 6-15 objects. Similarly, RT-1 [5] and RoboSet [2] also collect large-scale manipulation datasets but focus on a limited set of skills. Concurrently to our work, BridgeData V2 [36] collects a dataset with 13 skills across 24 environments. In this paper, we present a larger dataset with a wider range of skills and environments, with more comprehensive information. More importantly, all previous datasets put less emphasize on contact-rich manipulation. Our dataset focus more in this case and include the crucial force modality during manipulation.

*b) Zero/One-shot imitation learning:* The objective of training policies that can transfer to new tasks based on robot/human demonstrations is not new. Early works [33, 29, 15] focused on imitation learning using high-level states such as trajectories. Recently, researchers [14, 10, 42, 18, 41, 31, 30, 44, 16, 35, 4, 39, 8, 26, 20, 27] have started exploring raw-pixel inputs with the advancement of deep neural networks. Additionally, the requirement of demonstrations has been reduced by eliminating the need for actions. Recent approaches have explored various one-shot task descriptors, including images [18, 4], language [35, 26, 5, 2], robot video [14, 8, 27], or human video [41, 20]. These methods can be broadly classified into three categories: model-agnostic meta-learning [14, 41, 18, 4, 44], conditional behavior cloning [10, 8, 20, 5, 27], and task graph construction [16, 17]. While significant progress has been made in this direction, these approaches only consider visual observations and primarily focus on simple robotic manipulations such as reach, pick, push, or place. Our dataset offers the opportunity to take a step further by enabling the learning of *hundreds* of skills that require *multi-modal perception* within a single imitation learning model.

*c) Multi-Modal Learning of Vision and Force:* Force perception plays a crucial role in manipulation tasks, providing valuable and complementary information when visual perception is occluded. The joint modeling of vision and force in robotic manipulation has recently garnered interest within the research community [12, 25, 13, 24, 1, 7, 37]. However, most of these studies overlook the asynchronous nature of different modalities and simply concatenate the signals before or after the neural network. Moreover, the existing research primarily focuses on designing multi-modal learning algorithms for specific tasks, such as grasping [7], insertion [24], twisting [12], or playing Jenga [13]. A recent attempt [38] explores jointly imitating the action and wrench on 6 tasks respectively. Overall, the question of how to effectively handle multi-modal perception at different frequencies for various skills in a coherent manner remains open in robotics. Our dataset presents an opportunity for exploring multi-sensory learning across diverse real-world skills.

| Dataset | # Traj. | # Skills | # Robots | Human Demo | Contact Rich | Depth Sensing | Camera Calib. | Force Sensing |
|---|---|---|---|---|---|---|---|---|
| MIME [34] | 8.30k | 12 | 1 | ✔ | ✘ | ✔ | ✘ | ✘ |
| RoboTurk [28] | 2.10k | 2 | 1 | ✘ | ✘ | ✘ | ✘ | ✘ |
| RoboNet [9] | 162k | N/A | 7 | ✘ | ✘ | ✘ | ✘ | ✘ |
| BridgeData [11] | 7.20k | 4 | 1 | ✘ | ✘ | ✔* | ✘ | ✘ |
| BC-Z [20] | 26.0k | 3 | 1 | ✔ | ✘ | ✘ | ✘ | ✘ |
| RoboSet [2] | 98.5k | 12 | 1 | ✘ | ✔ | ✔ | ✘ | ✘ |
| BridgeData V2 [36] | 60.1k | 13 | 1 | ✘ | ✔ | ✔* | ✘ | ✘ |
| **RH20T** | 110k | 42 | 4 | ✔ | ✔ | ✔ | ✔ | ✔ |

TABLE I: Comparison with previous public datasets: "Camera Calib." indicates extrinsic calibration of all cameras and the robot. "✔*" indicates that only a portion of the images are paired with depth sensing. This comparison highlights the comprehensiveness of our dataset, which is the most extensive dataset for robotic manipulation to date.

| Conf. | Robot | Gripper | 6DoF F/T Sensor | Tactile |
|---|---|---|---|---|
| Cfg 1 | Flexiv | Dahuan AG95 | OptoForce | N/A |
| Cfg 2 | Flexiv | Dahuan AG95 | ATI Axia80-M20 | N/A |
| Cfg 3 | UR5 | WSG50 | ATI Axia80-M20 | N/A |
| Cfg 4 | UR5 | Robotiq-85 | ATI Axia80-M20 | N/A |
| Cfg 5 | Franka | Franka | Franka | N/A |
| Cfg 6 | Kuka | Robotiq-85 | ATI Axia80-M20 | N/A |
| Cfg 7 | Kuka | Robotiq-85 | ATI Axia80-M20 | uSkin |

TABLE II: Hardware specification of different configurations.

| Conf. | Modal | Size | Frequency |
|---|---|---|---|
| Cfg 1-7 | RGB image | 1280×720×3 | 10 Hz |
| | Depth image | 1280×720 | 10 Hz |
| | Binocular IR image | 1280×720 | 10 Hz |
| | Robot joint angle | 6 / 7 | 10 Hz |
| | Robot joint torque | 6 / 7 | 10 Hz |
| | Gripper Cartesion pose | 6 / 7 | 100 Hz |
| | Gripper width | 1 | 10 Hz |
| | 6DoF F/T | 6 | 100 Hz |
| | Audio | N/A | 30 Hz |
| Cfg 7 | Tactile | 2×16×3 | 200 Hz |

TABLE III: Data information of different configurations. The first 9 data modality are the same for all robot configurations. The last data modality of fingertip tactile sensing is only available in Cfg 7.

## III. RH20T Dataset

We introduce our robotic manipulation dataset, Robot-Human demonstration in 20TB (RH20T), to the community. Fig. 1 shows an overview of our dataset.

### A. Properties of RH20T

RH20T is designed with the objective of enabling general robotic manipulation, which means that the robot can perform various skills based on a task description, typically a human demonstration video, while minimizing the notion of rigid tasks. The following properties are emphasized to fulfill this objective, and Tab. I provides a comparison between our dataset and previous representative publicly available datasets.

*a) Diversity:* The diversity of RH20T encompasses multiple aspects. To ensure task diversity, we selected 48 tasks from RLBench [19], 29 tasks from MetaWorld [40], and introduced 70 self-proposed tasks that are frequently encountered and achievable by robots. In total, it contains 147 tasks, consisting of 42 skills (*i.e.,* verbs). Hundreds of objects were collected to accomplish these tasks. To ensure applicability across different robot configurations, we used 4 popular robot arms, 4 different robotic grippers, and 3 types of force-torque sensors, resulting in 7 robot configurations. Details about the robot configurations are provided in Tab. II.

To enhance environment diversity, we frequently replaced over 50 table covers with different textures and materials, and introduced irrelevant objects to create distractions. Manipulations were performed by tens of volunteers, ensuring diverse trajectories. To increase state diversity, for each skill, volunteers were asked to change the environmental conditions and repeat the manipulation 10 times, including variations in object instances, locations, and more. Additionally, we conducted robotic manipulation experiments involving human interference, both in adversarial and cooperative settings. Further details about each task are provided in the appendix.

*b) Multi-Modal:* We believe that the future of robotic manipulation lies in multi-modal approaches, particularly in open environments, where data from different sensors will become increasingly accessible with advancements in technology. In the current version of RH20T, we provide visual, tactile, audio, and proprioception information. Visual perception includes RGB, depth, and binocular IR images from three types of cameras. Tactile perception includes 6 DoF force-torque measurements at the robot's wrist, and some sequences also include fingertip tactile information. Audio data includes recordings from both in-hand and global sources. Proprioception encompasses joint angles/torques, end-effector Cartesian pose and gripper states. All information is collected at the highest frequency supported by our workstation and saved with corresponding timestamps, and the details are given in Tab. III.

*c) Scale:* Our dataset consists of over 110,000 robot sequences and an equal number of human sequences, with more than 50 million images collected in total. On average, each skill contains approximately 750 robot manipulations. Fig. 2 provides a detailed breakdown of the number of manipulations across different tasks in the dataset, showing a relatively uniform distribution. Fig. 3 presents statistics on the manipulation time for each sequence in our dataset. Most sequences have durations ranging from 10 to 100 seconds. With its substantial volume of data, our dataset stands as the largest in our community at present.

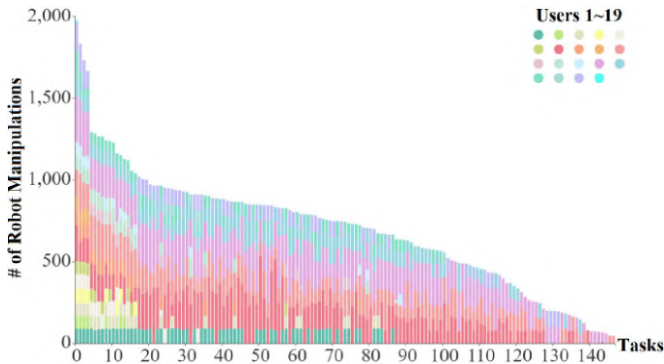

Fig. 2: Statistics on the amount of robotic manipulation for different tasks.

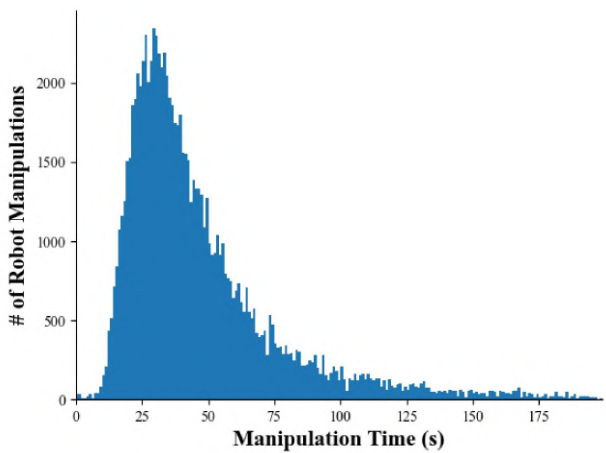

Fig. 3: Statistics on the execution time of different robotic manipulations in our dataset.

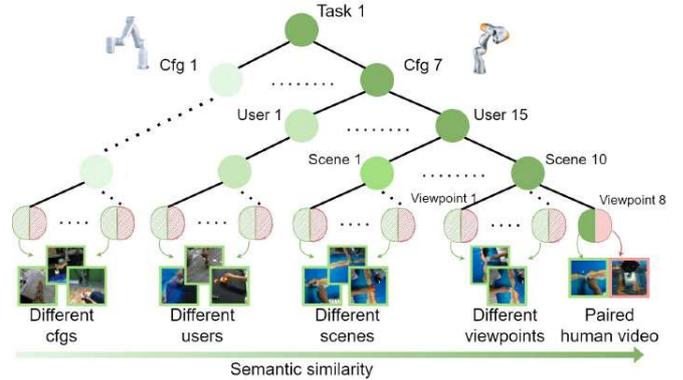

Fig. 4: Example of data hierarchy: The leaf nodes in the hierarchy consist of human demonstrations (highlighted in green) and robot manipulations (highlighted in red, only the right-est example is shown in the figure). We can pair a robot manipulation sequence with human demonstration videos captured from different viewpoints, scenes, human subjects, and environments. Zoom in to explore the details of various human demonstrations.

*d) Data Hierarchy:* Humans can accurately understand the semantics of a task based on visual observations, regardless of the viewpoint, background, manipulation subject, or object. We aim to provide a dataset that offers dense <human demonstration, robot manipulation> pairs, enabling models to learn this property. To achieve this, we organize the dataset in a tree hierarchy based on intra-task similarity. Fig. 4 illustrates an example tree structure and the criteria at different levels. Leaf nodes with a more recent common ancestor are more closely related. For each task, millions of <human demonstration, robot manipulation> pairs can be constructed by pairing leaf nodes with a common ancestor at different levels.

*e) Compositionality:* RH20T includes not only short sequences that perform single manipulations but also long manipulation sequences that combine multiple short tasks. For example, a sequence of actions such as grabbing the plug, plugging it into the socket, turning on the socket switch, and turning on the lamp can be considered as a single task, with each step also being a task. This task composition allows us to investigate whether mastering short sequences improves the acquisition of long sequence tasks.

## B. Data Collection and Processing

Unlike previous methods that simplify the tele-operation interface using 3D mice, VR remotes, or mobile phones, we place emphasis on the importance of intuitive and accurate tele-operation in collecting contact-rich robot manipulation data. Without proper tele-operation, the robot could easily collide with the environment and generate significant forces, triggering emergency stops. Consequently, previous works either avoid contact [20] or operate at reduced speeds to mitigate these risks.

*a) Collection:* Fig. 5 shows an example of our data collection platform. Each platform contains a robot arm with force-torque sensor, gripper and 1-2 inhand cameras, 8-10 global cameras, 2 microphones, a haptic device, a pedal and a data collection workstation. All the cameras are extrinsically calibrated before conducting the manipulation. The human demonstration video is collected on the same platform by human with an extra ego-centric camera. Tens of volunteers conducted the robotic manipulation according to our task lists and text description. We make our tele-operation pretty intuitive and the average training time is less than 1 hour. The volunteers are also required to specify ending time of the task and give a rating from 0 to 9 after finishing each manipulation. 0 denotes the robot enters the emergency state (e.g., hard collision), 1 denotes the task fails and 2-9 denotes their evaluation of the manipulation quality. The success and failure cases have a ratio of around 10:1 in our dataset.

*b) Processing:* We preprocess the dataset to provide a coherent data interface. The coordinate frame of all robots and force-torque sensors are aligned. Different force-torque sensors are tared carefully. The end-effector Cartesian pose and the force-torque data are transformed into the coordination system of each camera. Manual validation is performed for each scene to ensure the camera calibration quality. Fig. 6 shows an illustration of rendering different component of the data in a unified coordinate frame and demonstrates the high-quality of our dataset. The detailed data format and data access APIs are provided on our website.

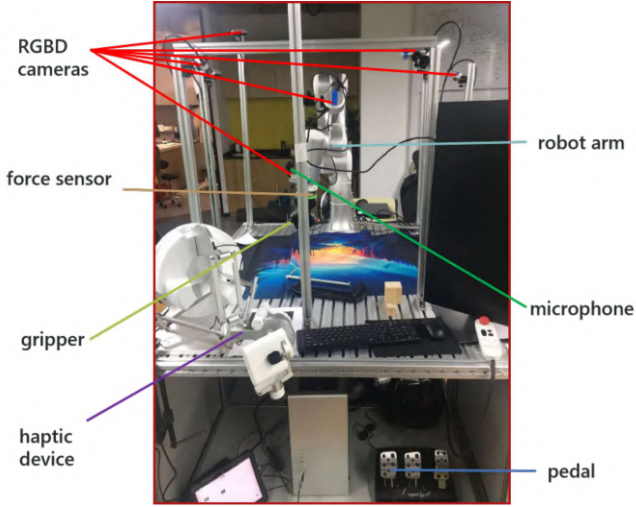

Fig. 5: Illustration of our data collection platform

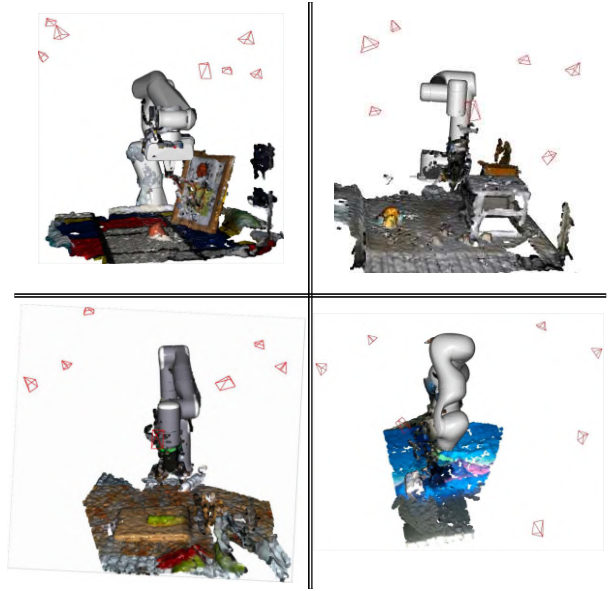

Fig. 6: We display the point cloud generated by fusing the RGBD data from the multi-view cameras mounted in our data collection platform. The red pyramids indicate the camera poses. Additionally, the robot model is rendered in the scene based on the joint angles recorded in our dataset. It is evident that all the cameras are calibrated with respect to the robot's base frame, and all the recorded data are synchronized in the temporal domain.

## IV. Experiments

We introduce the RH20T dataset in pursuit of enabling robots to acquire novel skills within unfamiliar environments using minimal data. While the ultimate objective is to train a large model capable of performing such tasks in a one-shot learning fashion, we acknowledge the significant computational resources required for this endeavor, which are presently beyond our reach. Consequently, this paper primarily focuses on demonstrating the dataset's effectiveness in enhancing the transferability of a baseline model within a few-shot learning framework.

To assess the efficacy of our dataset, we adopt the Action Chunking with Transformers (ACT) model as our baseline network. ACT, as proposed by in a recent work [43], has demonstrated remarkable capabilities in handling complex robot manipulation tasks. It leverages the power of transformers to learn intricate action sequences from hundreds of demonstrations.

### A. Experimental Setup

*a) Platform:* In our experiments, we utilize a Flexiv robot arm equipped with an Intel RealSense RGB-D camera in front of the robot for perceiving the environment and a Dahuan-95 gripper for interacting with objects. We set up a new environment where the camera pose and table cover are different from those in our RH20T dataset. Fig. 7 (a) illustrates our robot platform.

*b) Procedure:* We setup a task involves grasping a block and placing it on a weight. In the new environment, we collect 75 robotic manipulation sequences, including RGB images and actions, through teleoperation. From our dataset, we select 335 robotic manipulation from the same task and 195 manipulation from 3 different but similar tasks (pick up a block; pick up a block and place it at the designated location; pick up a block and move it from left to right). All the manipulation sequences from our dataset have different camera views, table covers, objects and robot embodiments from the robotic environment in our current experiment.

We initiate the training process by pre-training the ACT model on different subsets of the data selected from our dataset. By exposing the model to a range of robotic manipulation scenarios, we aim to enhance its ability to generalize across various tasks and environmental conditions. Following pre-training, we fine-tune the ACT model on specific portions of the newly collected data, focusing on the task involving grasping and weight placement. This stage aims to refine the model's performance on the target task.

We evaluate the performance of the ACT model both with and without pre-training on our dataset. The experiments are carried out on the real robot platform and repeated for 20 times for each configuration. We divide the task into 3 stages, namely whether the robot can reach the block, grasp it and place it on the weight, and measure the success rate at each stage. Additionally, we examine how well the model generalizes to variations in object properties. The evaluation time limit is set as 60 seconds.

*c) Implementation Details:* For ACT model, we set the hidden channel and the feedforward channel in the network to 512 and 3200 respectively. During pre-training phase, the model is trained with a learning rate of $2 \times 10^{-5}$ for 10 epochs; while during fine-tuning phase, the model is trained with a learning rate of $10^{-5}$ for 750 epochs. Although it is less than the original implementation [43], we increase the sample density per epoch by including all valid sub-trajectories of the newly collected demonstrations. Hence,

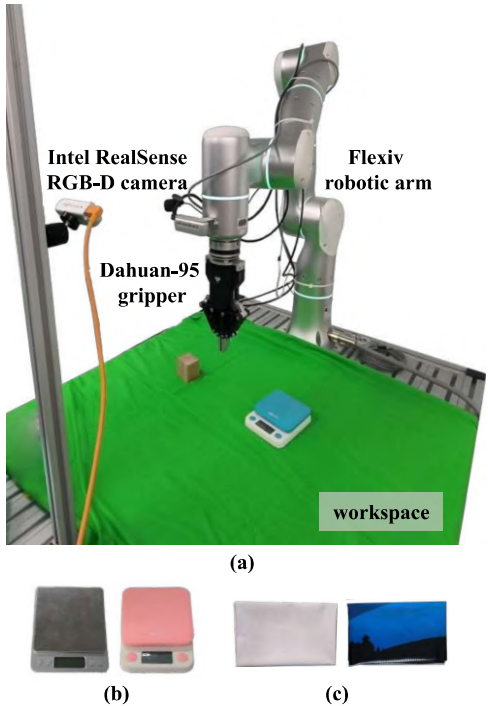

Fig. 7: (a) The experimental robot platform. (b) Varied weights (metal, pink) assessing the model's generalization ability. (c) Distinct table covers (white, blue) evaluating the model's generalization ability.

| # Demos | Pretrain Same | Task Multi. | Training Epochs | Success Rate (%) ↑ Reach | Pick | Place |
|---|---|---|---|---|---|---|
| 75 |  |  | 500 | 35 | 10 | 0 |
|  | ✓ |  | 500 | 70 | 15 | 15 |
|  | ✓ | ✓ | 500 | 65 | 20 | 15 |
|  |  |  | 750 | 55 | 5 | 0 |
|  | ✓ |  | 750 | **80** | 20 | 15 |
|  | ✓ | ✓ | 750 | **80** | 25 | 25 |
| 40 |  |  | 750 | 45 | 10 | 0 |
|  | ✓ |  | 750 | 65 | **25** | 5 |
|  | ✓ | ✓ | 750 | 70 | **25** | 15 |
| 10 |  |  | 750 | 15 | 0 | 0 |
|  | ✓ |  | 750 | 30 | **15** | 5 |
|  | ✓ | ✓ | 750 | **50** | 10 | 5 |

TABLE IV: Experimental results of ACT trained in different settings and tested in the original environment (20 trials).

| Type |  | w/wo pretrain | Success Rate (%) ↑ Reach | Pick | Place |
|---|---|---|---|---|---|
| weight | metal |  | 20 | 0 | 0 |
|  |  | ✓ | **80** | **30** | **10** |
|  | pink |  | 40 | 10 | 0 |
|  |  | ✓ | **70** | **20** | **10** |
| table cover | white |  | 20 | 0 | 0 |
|  |  | ✓ | **50** | 0 | 0 |
|  | blue |  | 30 | 0 | 0 |
|  |  | ✓ | **80** | **20** | **10** |

TABLE V: Experimental results of ACT trained in different settings and tested in different environments (10 trials).

750 epochs are sufficient for the model to converge well. The chunk size is set to 20, which corresponds to 2 seconds with the frequency of 10Hz. The images are scaled to $640 \times 360$ during training and testing. We apply temporal ensembling and set its coefficient $k = 0.01$ following [43] in evaluation.

### B. Experimental Results

We present the model's success rates under different training configurations in Tab. IV. When training the network with 75 demonstrations, we observe that pretraining the model with selected data from our dataset, despite differences in camera viewpoints, robot embodiments, and backgrounds, enhances the final success rate. Additionally, the inclusion of data from different tasks during pretraining further improves the overall success rate. Comparing the results of training for 500 epochs with pretraining to training for 750 epochs without pretraining, we find that pretraining on our dataset also accelerates model convergence. These results demonstrate that leveraging the diverse training data from our dataset enhances the adaptability and robustness of the robotic manipulation model.

We then reduce the number of demonstrations collected in this new environment to simulate a few-shot learning scenario. With 40 robot demonstrations, the results of pretraining on our dataset outperform the counterpart trained with 75 demonstrations without pretraining. Further reducing the demonstrations to 10, the results of pretraining on multiple tasks from our dataset still surpass the one trained with 75 demonstrations without pretraining. This demonstrates the beneficial impact of our dataset on few-shot learning in robotic manipulation.

Finally, we replace the object and table cover used during testing with novel ones to assess the models' generalization ability in new environments. The weights and table covers used for replacement are shown in Fig. 7(b) and (c). In this experiment, we compare two models, both are trained with the original 75 demonstrations for 750 epochs, one with pretraining on multiple similar tasks from our dataset and one without. The experimental results in Tab. V demonstrate that the model pretrained on our dataset consistently outperforms its counterpart without pretraining, indicating that our dataset enhances the model's generalization ability.

## V. DISCUSSION AND CONCLUSION

In this paper we present the RH20T dataset for diverse robotic skill learning. We believe it can facilitate many areas in robotics, especially for robotic manipulation in novel environments. The current limitations of this paper are that (i) the cost of data collection is expensive and (ii) the potential of robotic foundation models is not evaluated on our dataseet. We have tried to duplicate the results of some recent robotic foundation models but haven't succeeded yet due the limit of computing resources. Thus, we decide to open source the dataset at this stage and hope to promote the development of this area together with our community. In the future, we hope to extend our dataset to broader robotic manipulation, including dual-arm and multi-finger dexterous manipulation.

*Author contributions:* H.-S. Fang initiated the project, set up the robot platform, initialized the tele-operation toolkit, curated the data collection pipeline, and wrote the paper. H. Fang set up the robot platform, optimized the tele-operation toolkit, assisted with policy training, and wrote the project page. Z. Tang assisted with data collection, calibrated the sensors, structured the dataset, and wrote the data access API. C. Wang trained the policy. J. Liu explored one-shot imitation learning with transformer architecture. J. Wang assisted with data collection and dataset parsing. H. Zhu explored annotating human keypoints for the human demonstration video. C. Lu supervised the project and provided hardware and resource support.

# Appendix Task Specification of RH20T

Table 1: Task description for our dataset. "Src." denotes the source of the task. Note that the task IDs are not necessarily continuous.

| Items | Task Desc. | Src. | Items | Task Desc. | Src. | Items | Task Desc. | Src. |
|---|---|---|---|---|---|---|---|---|
| 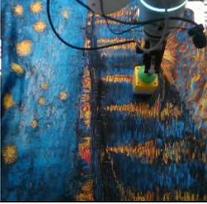 | 1. Press the button from top to bottom | Meta-World | 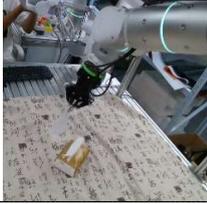 | 2. Pull out a napkin | Self-Proposed | 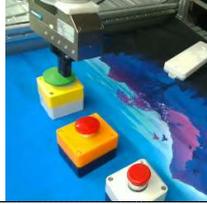 | 3. Press three buttons from left to right in sequence | RLBench |
| 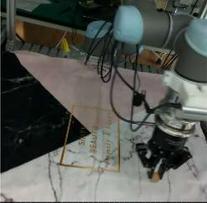 | 4. Pick up a block on the left and move it to the right | Meta-World | 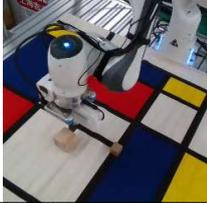 | 5. Approach and touch the side of a block | Meta-World | 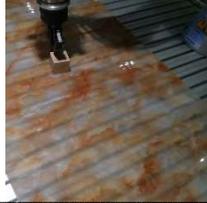 | 6. Use the gripper to push a block from left to right | Meta-World |
| 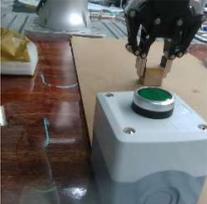 | 7. Hold a block with the gripper and sweep it from left to right on the table | Meta-World | 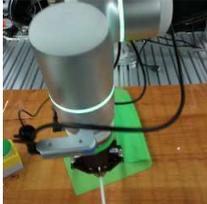 | 8. Grab a block and place it at the designated location | RLBench | 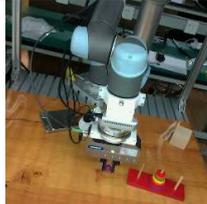 | 9. Take out one Hanoi block and throw it aside | RLBench |
| 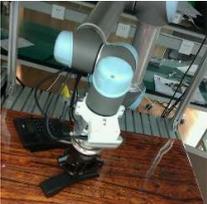 | 10. Place the handset of the telephone on the corresponding phone cradle | RLBench | 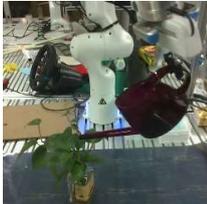 | 11. Water the plant | RLBench | 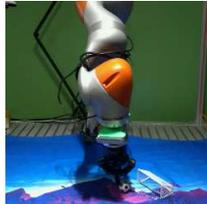 | 12. Push the soccer ball into the goal | Meta-World |
| 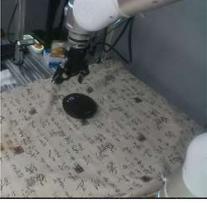 | 13. Place the block on the scale | RLBench | 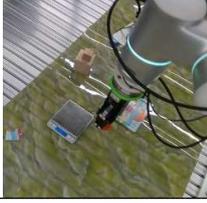 | 14. Remove the object from the scale | RLBench | 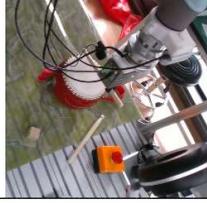 | 15. Play the drum | Self-Proposed |
| 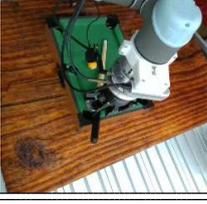 | 16. Hit the pool ball | RLBench | 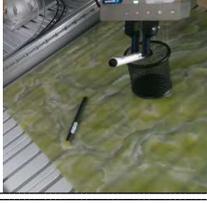 | 17. Put the pen into the pen holder | RLBench | 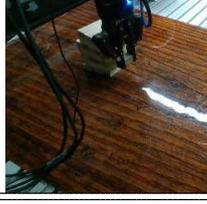 | 18. Play Jenga | RLBench |

| Items | Task Desc. | Src. | Items | Task Desc. | Src. | Items | Task Desc. | Src. |
|---|---|---|---|---|---|---|---|---|
| 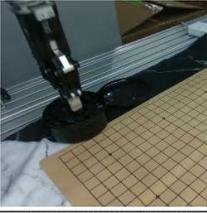 | 19. Play the first move as black in the upper right corner of the Go board | Self-Proposed | 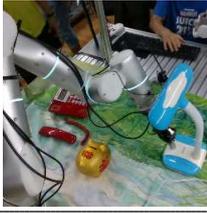 | 20. Turn on the desk lamp by pressing the button | RLBench | 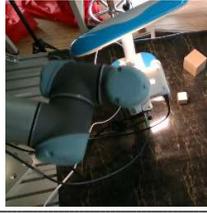 | 21. Turn off the desk lamp by pressing the button | RLBench |
| 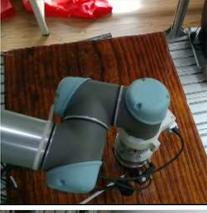 | 22. Wave the flag | Self-Proposed | 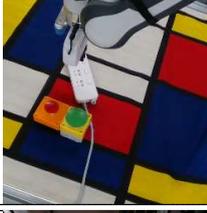 | 23. Turn on the power strip by pressing the button | Self-Proposed | 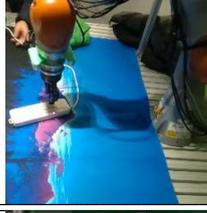 | 24. Turn off the power strip by pressing the button | Self-Proposed |
| 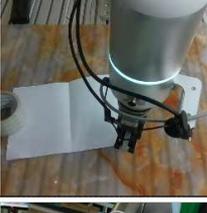 | 25. Unfold a piece of paper | Self-Proposed | 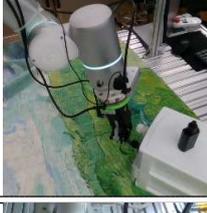 | 26. Use the gripper to push and close the drawer | Meta-World | 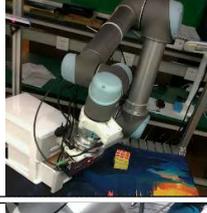 | 28. Grasp the handle and close the drawer | RLBench |
| 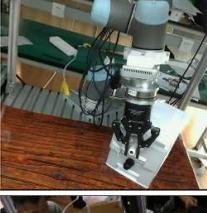 | 29. Grasp the handle and open the drawer | RLBench | 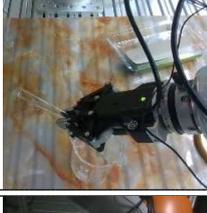 | 30. Pour out the test tube | Self-Proposed | 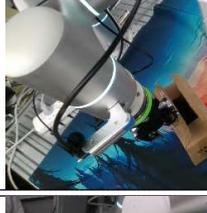 | 31. Cover the box | Meta-World |
| 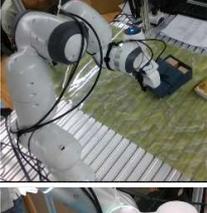 | 32. Slide the outer casing onto the gift box | Self-Proposed | 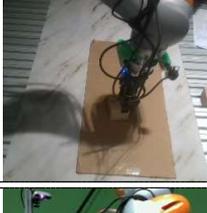 | 33. Grasp one block to sweep the other block onto the mark | Meta-World | 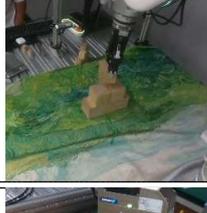 | 34. Stack the squares into a pyramid shape | RLBench |
| 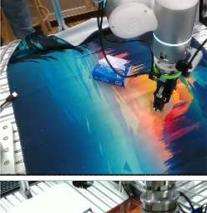 | 35. Pick up one small block | RLBench | 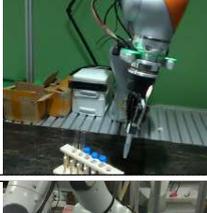 | 36. Shake the test tube | Self-Proposed | 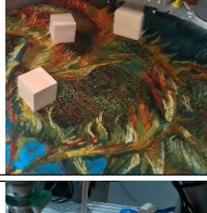 | 37. Stack the blocks in a vertical line of five | RLBench |
| 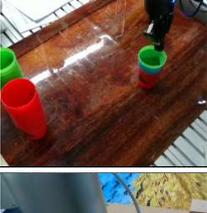 | 38. Pick up the cup | RLBench | 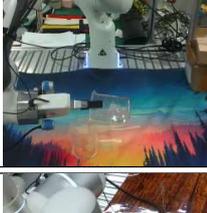 | 39. Pour the water from one cup into another empty cup | blabla | 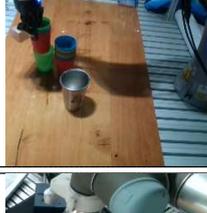 | 40. Stack the cups | RLBench |
| 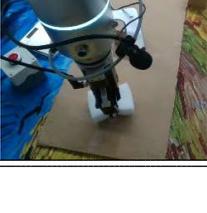 | 41. Clean the tabletop with a sponge | RLBench | 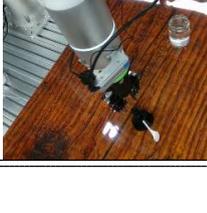 | 42. Screw the lid onto the jar | RLBench | 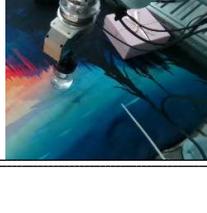 | 43. Unscrew the lid from the jar | RLBench |

| Items | Task Desc. | Src. | Items | Task Desc. | Src. | Items | Task Desc. | Src. |
|---|---|---|---|---|---|---|---|---|
| 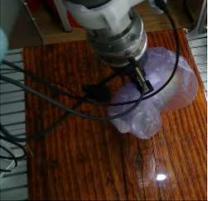 | 44. Pick up a bag of things | Self-Proposed | 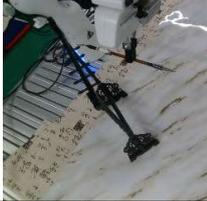 | 45. Hang the brush on the pen rack | Self-Proposed | 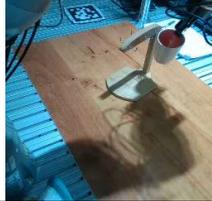 | 46. Hang the cup on the cup rack | RLBench |
| 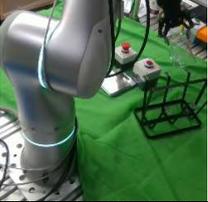 | 47. Take the cup off the cup rack | RLBench | 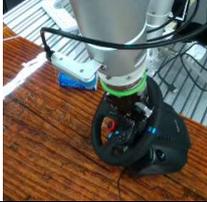 | 48. Rotate the steering wheel 90 degrees clockwise | Self-Proposed | 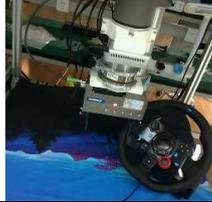 | 49. Rotate the steering wheel 90 degrees counter-clockwise | Self-Proposed |
| 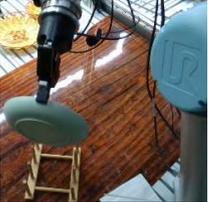 | 50. Put the dish on the dish rack | Self-Proposed | 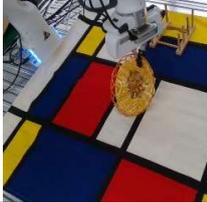 | 51. Take the dish off the dish rack | Self-Proposed | 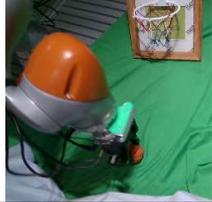 | 52. Grab a basketball, release it and shoot it into the basket | Meta-World |
| 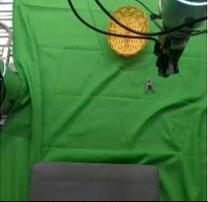 | 53. Use a clamp | Meta-World | 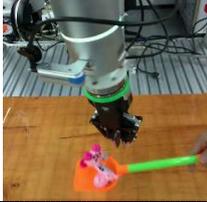 | 54. Catch the moving object | Self-Proposed | 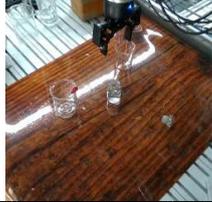 | 55. Transfer liquid using a dropper | Self-Proposed |
| 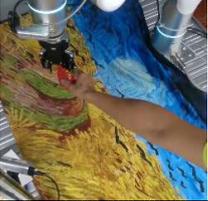 | 56. Receive something handed over by a human | Self-Proposed | 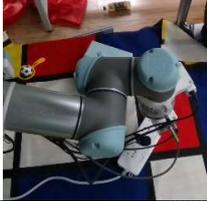 | 57. Turn on the four buttons on the power strip | Self-Proposed | 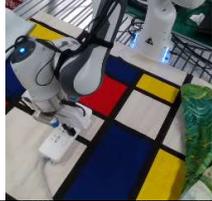 | 58. Turn off the four buttons on the power strip | Self-Proposed |
| 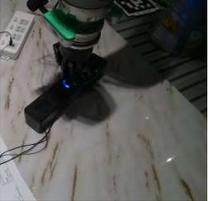 | 59. Turn the knob to increase the volume of a speaker | Self-Proposed | 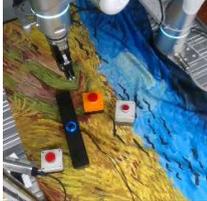 | 60. Turn the knob to decrease the volume of the speaker | Self-Proposed | 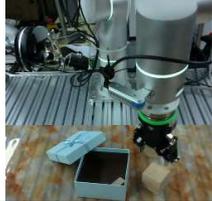 | 61. Take everything out of the gift box | Self-Proposed |
| 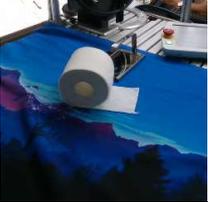 | 62. Put the toilet paper on its holder | Self-Proposed | 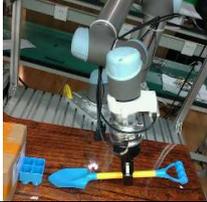 | 63. Use a shovel to scoop up an object | Self-Proposed | 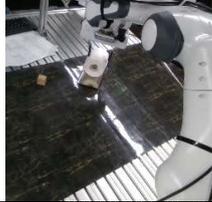 | 64. Take the toilet paper off its holder | Self-Proposed |
| 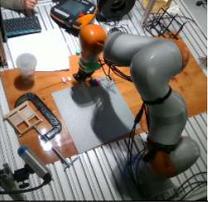 | 65. Build with small Lego blocks | Self-Proposed | 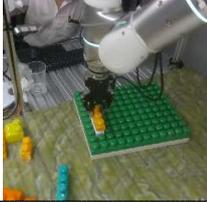 | 66. Build with large Megabloks | Self-Proposed | 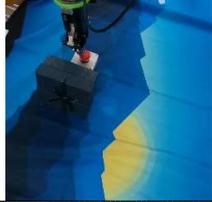 | 67. Press a button from top to bottom with obstacles | Meta-World |

| Items | Task Desc. | Src. | Items | Task Desc. | Src. | Items | Task Desc. | Src. |
|---|---|---|---|---|---|---|---|---|
| 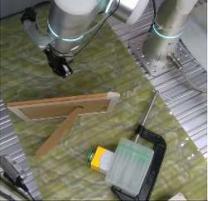 | 68. Press a button horizontally with obstacles | Meta-World | 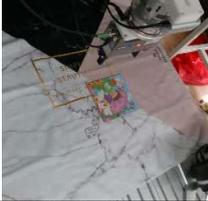 | 69. Assemble one piece of a puzzle | RLBench | 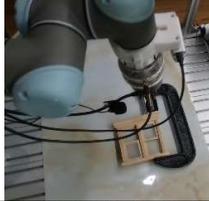 | 70. Open a sliding window | Meta-World |
| 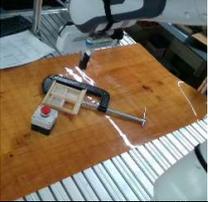 | 71. Close a sliding window | Meta-World | 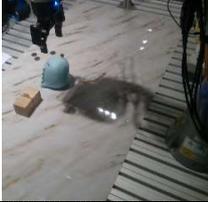 | 72. Drop coins into a piggy bank | Self-Proposed | 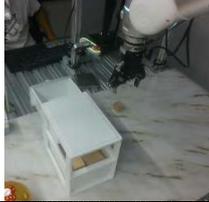 | 73. Put things in the drawer | RLBench |
| 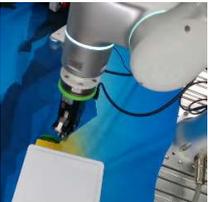 | 74. Press the button horizontally | Meta-World | 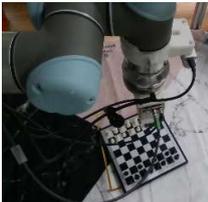 | 75. Finish setting up the starting position of a chessboard that is almost arranged | Self-Proposed | 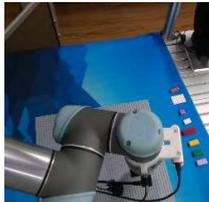 | 76. Stack blocks (small Lego) one on top of the other every time | Self-Proposed |
| 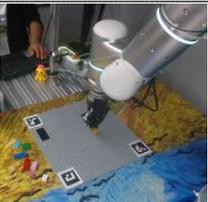 | 77. Stack blocks (small Lego) randomly one at a time | Self-Proposed | 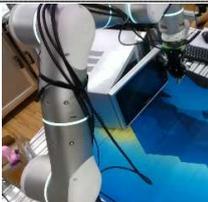 | 78. Close the microwave door | RLBench | 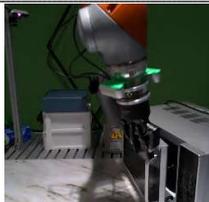 | 79. Open the microwave door | RLBench |
| 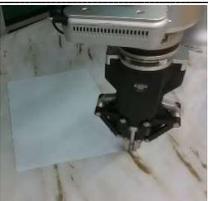 | 80. Flip over and spread out the paper that is laid flat on the table | Self-Proposed | 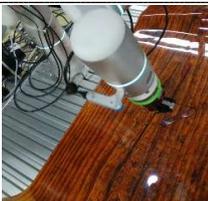 | 81. Unfold the leg of the glasses (with one hand) | Self-Proposed | 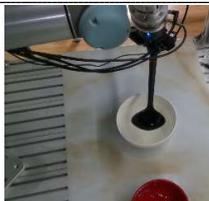 | 82. Scoop water with a large spoon from one bowl to another | Self-Proposed |
| 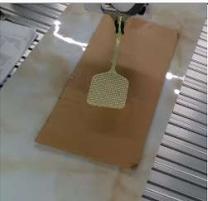 | 83. Swat with a flyswatter | Self-Proposed | 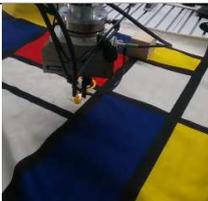 | 84. Assemble: Attach the bubble ring to the ball | Meta-World | 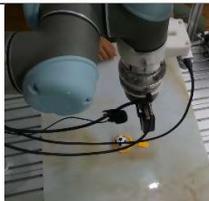 | 85. Remove the bubble ring from the assembled bubble ring and ball | Meta-World |
| 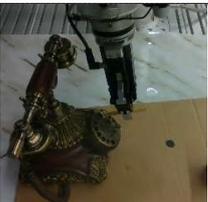 | 86. Dial a number on an old rotary phone | Meta-World | 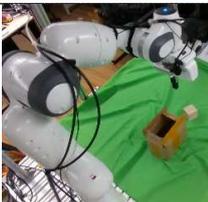 | 88. Pick up and place an object with obstacles | Meta-World | 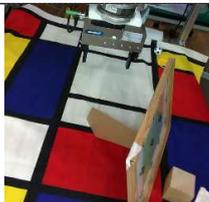 | 89. Push an object with obstacles | Meta-World |
| 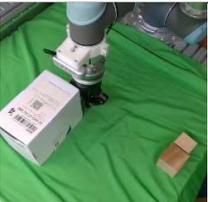 | 90. Approach and touch an object with obstacles | Meta-World | 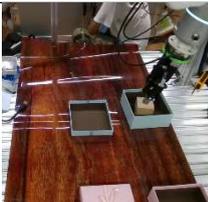 | 91. Move an object from one box to another | Meta-World | 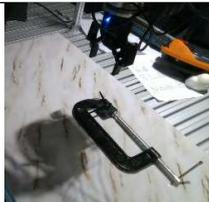 | 92. Turn the hands of a clock | RLBench |

| Items | Task Desc. | Src. | Items | Task Desc. | Src. | Items | Task Desc. | Src. |
|---|---|---|---|---|---|---|---|---|
| 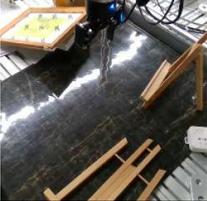 | 93. Put the photo frame on the bracket | RLBench | 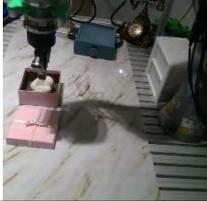 | 94. Open a box | RLBench | 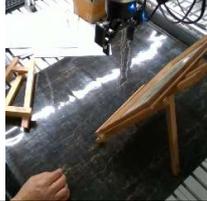 | 95. Take the photo frame down from the bracket | RLBench |
| 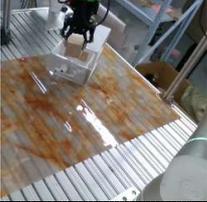 | 96. Take something out of a drawer | RLBench | 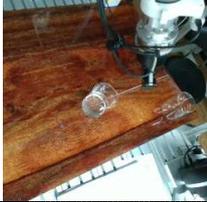 | 100. Stir the beaker with a glass rod | Self-Proposed | 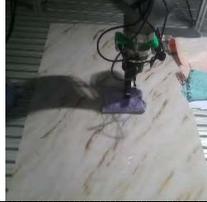 | 101. Clean the table with a cloth | Self-Proposed |
| 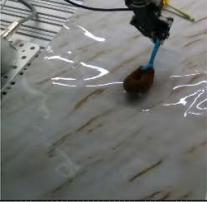 | 102. Scrub the table with a brush | Self-Proposed | 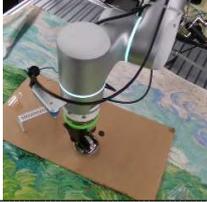 | 103. Drag the plate to the goal post after holding it down | Meta-World | 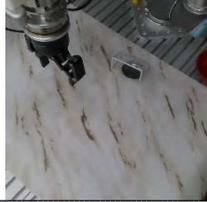 | 104. Drag the plate back after holding it down | Meta-World |
| 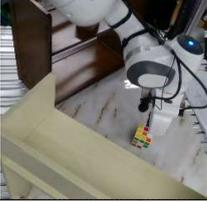 | 105. Put the object on the shelf | Meta-World | 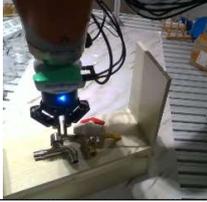 | 106. Take the object down from the shelf | Meta-World | 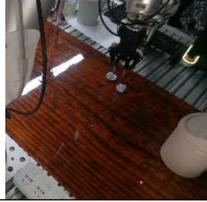 | 107. Put the garbage in the trash can | RLBench |
| 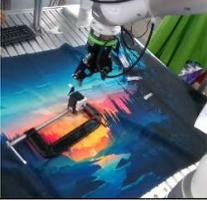 | 108. Sharpen the pencil with a pencil sharpener | Self-Proposed | 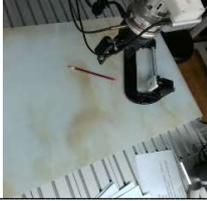 | 109. Insert the pencil into the pencil sharpener | Self-Proposed | 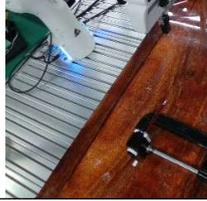 | 110. Take the pencil out from the pencil sharpener | Self-Proposed |
| 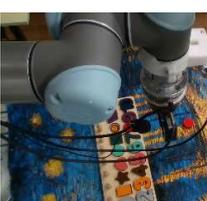 | 111. Put the object with the corresponding shape into the corresponding hole | RLBench | 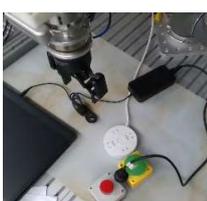 | 112. Plug in the charger to the socket | Self-Proposed | 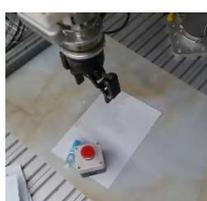 | 116. Use the correction tape on paper | Self-Proposed |
| 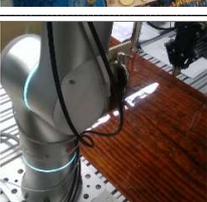 | 118. Turn on the water tap | Meta-World | 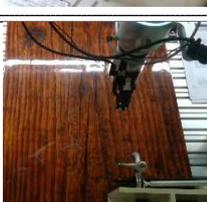 | 119. Turn off the water tap | Meta-World | 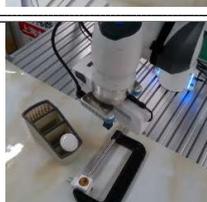 | 120. Install the light bulb by rotating it | RLBench |
| 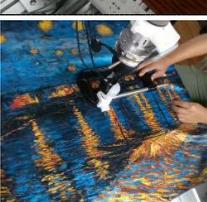 | 121. Take out the light bulb by rotating it | RLBench | 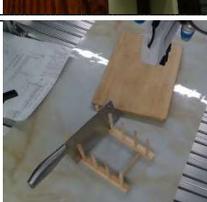 | 122. Put the knife on the cutting board | RLBench | 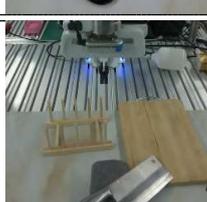 | 123. Put the knife on the knife rack | RLBench |

| Items | Task Desc. | Src. | Items | Task Desc. | Src. | Items | Task Desc. | Src. |
|---|---|---|---|---|---|---|---|---|
| 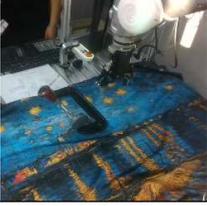 | 124. Push down the lever | Meta-World | 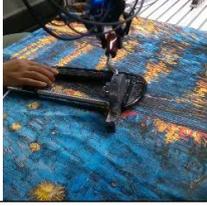 | 125. Pull up the lever | Meta-World | 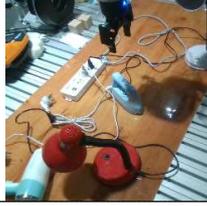 | 126. Plug in the power cord to the socket | Self-Proposed |
| 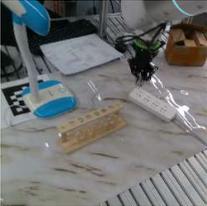 | 127. Plug in the power cord of the desk lamp, turn on the socket, and light up the desk lamp | Self-Proposed | 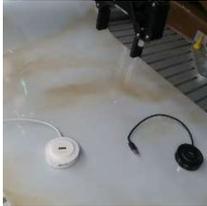 | 128. Plug in the USB drive to the docking station | RLBench | 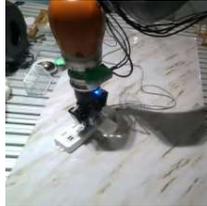 | 129. Plug in the bulb holder with a bulb to the socket | Self-Proposed |
| 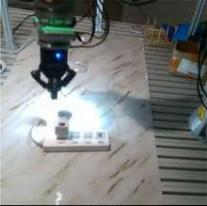 | 130. Plug in the bulb holder with a bulb to the socket and turn on the switch of the bulb | Self-Proposed | 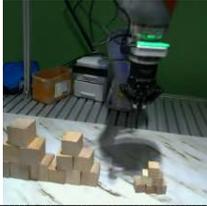 | 131. Stack the blocks into a pyramid | Self-Proposed | 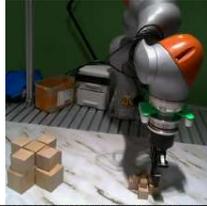 | 132. Stack the blocks into a cross shape | Self-Proposed |
| 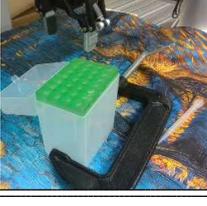 | 200. Insert the tip of a large pipette into the holder for large pipette tips | Self-Proposed | 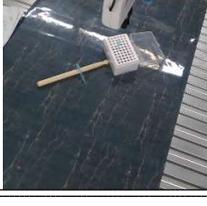 | 201. Insert the tip of a medium pipette into the holder for medium pipette tips | Self-Proposed | 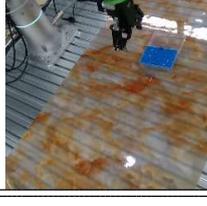 | 202. Insert the tip of a small pipette into the holder for small pipette tips | Self-Proposed |
| 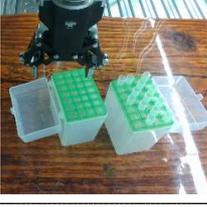 | 204. Transfer all large pipette tips from one holder to another holder for large pipette tips | Self-Proposed | 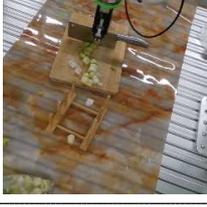 | 205. Chop the scallions | Self-Proposed | 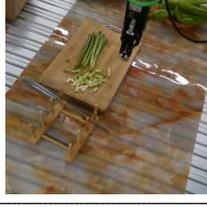 | 206. Chop the green garlic | Self-Proposed |
| 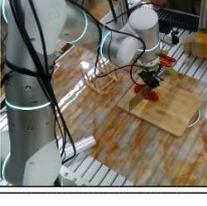 | 207. Chop the chili peppers | Self-Proposed | 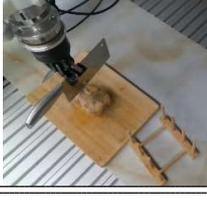 | 208. Slice the lotus root | Self-Proposed | 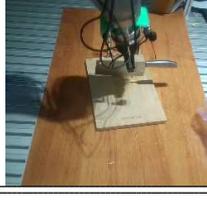 | 209. Slice the carrots | Self-Proposed |
| 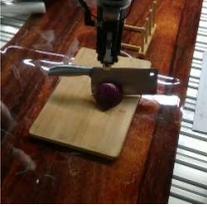 | 210. Chop the onions | Self-Proposed | 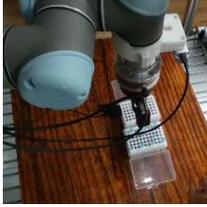 | 211. Transfer all medium pipette tips from one rack to another holder for medium pipette tips | Self-Proposed | 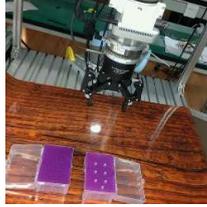 | 212. Transfer all small pipette tips from one rack to another holder for small pipette tips | Self-Proposed |
| 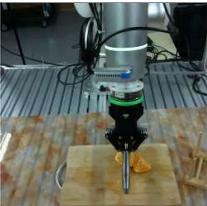 | 213. Chop the orange | Self-Proposed | 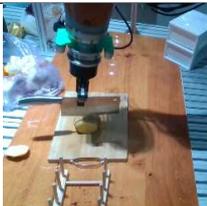 | 215. Chop the potatoes | Self-Proposed | 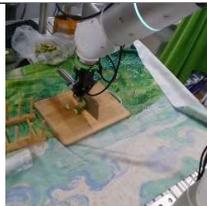 | 216. Chop the cucumber into shreds | Self-Proposed |

| Items | Task Desc. | Src. | Items | Task Desc. | Src. | Items | Task Desc. | Src. |
|---|---|---|---|---|---|---|---|---|
| 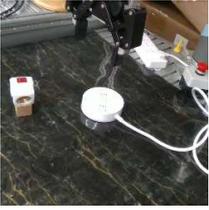 | 217. Plug in the bulb holder to the socket | Self-Proposed | 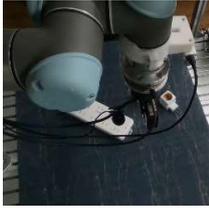 | 218. Plug in the bulb holder to the socket, install the bulb, turn on the socket to light up the bulb | Self-Proposed | 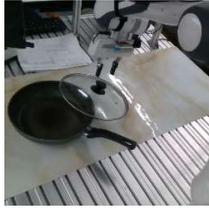 | 222. Cover the pot with the lid | RLBench |
| 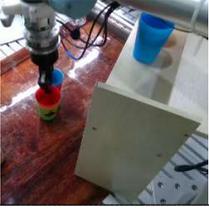 | 223. Take the cups off the shelf and stack them together | Self-Proposed | 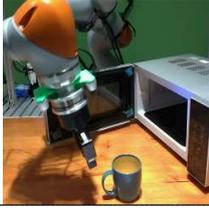 | 225. Put the bowl into the microwave | Self-Proposed | 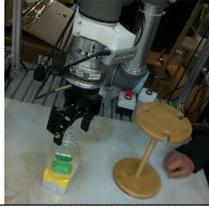 | 329. Put the glass cup onto the shelf | Self-Proposed |